# HiGNN: Hierarchical Informative Graph Neural Networks for Molecular Property Prediction Equipped with Feature-Wise Attention

Weimin Zhu,[†] Yi Zhang,[†] DuanCheng Zhao,[†] Jianrong Xu[‡,§] and Ling Wang*,[†]

[†]Guangdong Provincial Key Laboratory of Fermentation and Enzyme Engineering, Joint International Research Laboratory of Synthetic Biology and Medicine, Guangdong Provincial [‡]Engineering and Technology Research Center of Biopharmaceuticals, School of Biology and Biological Engineering, South China University of Technology, Guangzhou 510006, China.

[§]Department of Pharmacology and Chemical Biology, Shanghai Jiao Tong University School of Medicine, Shanghai, 200025, China.

**ABSTRACT:** Elucidating and accurately predicting the druggability and bioactivities of molecules plays a pivotal role in drug design and discovery and remains an open challenge. Recently, graph neural networks (GNN) have made remarkable advancements in graph-based molecular property prediction. However, current graph-based deep learning methods neglect the hierarchical information of molecules and the relationships between feature channels. In this study, we propose a well-designed hierarchical informative graph neural networks framework (termed HiGNN) for predicting molecular property by utilizing a co-representation learning of molecular graphs and chemically synthesizable BRICS fragments. Furthermore, a plug-and-play feature-wise attention block is first designed in HiGNN architecture to adaptively recalibrate atomic features after the message passing phase. Extensive experiments demonstrate that HiGNN achieves state-of-the-art predictive performance on many challenging drug discovery-associated benchmark datasets. In addition, we devise a molecule-fragment similarity mechanism to comprehensively investigate the interpretability of HiGNN model at the subgraph level, indicating that HiGNN as a powerful deep learning tool can help chemists and pharmacists identify the key components of molecules for designing better molecules with desired properties or functions. The source code is publicly available at https://github.com/idruglab/hignn.

## INTRODUCTION

Molecular property prediction is a vital but challenging task in pharmaceutical industrial workflows.[1] Accurate learning of the bioactive, physicochemical, and ADMET (absorption, distribution, metabolism, excretion, and toxicity) properties of molecules can substantially accelerate drug discovery and development by minimizing the time and cost needed for experimental high-throughput screening (HTS). Traditionally, computer-aided drug design (CADD) methods such as molecular docking,[2] molecular similarity,[3,4] pharmacophore modeling,[5] and quantitative structure-activity relationships (QSAR)[6] have been effective at narrowing screening compounds collections, and assist in ranking the druggability of compounds in experimental trials. These approaches usually develop prediction models based on a high level of domain-specific expertise. With the continuous increase in the amount of experimental data related to pharmacy, chemistry, and biology,[7,8] and the development of artificial intelligence (AI) techniques,[9] AI-aided drug design (AIDD), an innovative paradigm of computational method relying on big data, has attracted much attention in the past few decades and holds promising opportunities for future improvement of interdisciplinary studies.[10-12]

Typically, AI-based models endeavor to predict molecular properties using various molecular representations, such as molecular descriptors,[13] molecular fingerprints,[14] simplified molecular-input line-entry systems (SMILES),[15] molecular images, and molecular graphs. Conventional machine learning (CML) methods, such as support vector machine (SVM),[16,17] random forest (RF),[18,19] and extreme gradient boosting (XGBoost),[20,21] rely on the selection of hand-crafted molecular descriptors or molecular fingerprints, which suffer from complicated and time-consuming feature engineering. Recently, with the remarkable breakthroughs of deep learning (DL) methods in numerous fields such as computer vision (CV) and natural language processing (NLP),[22-24] such DL methods have been applied in the field of drug design and discovery. For example, DL methods have demonstrated powerful capabilities to automatically learn molecular representations from molecular structures, which further improve the performance of molecular property prediction. Deep neural networks (DNN), a prototypical DL architecture, was originally utilized for learning descriptor-based representations,[25] but it is not suitable for all kinds of raw molecular characteristics. Image-based[26,27] and sequence-based[28-30] methods for molecular property prediction have truly blossomed in the past few years. Whether utilizing convolutional neural networks (CNN) to process regular compound images information, or leveraging recurrent neural networks (RNN) and Transformer architectures[24] to operate on semantic molecular notations like SMILES, existing studies showed that these DL networks have led to tremendous advances in cheminformatics, computational chemistry, as well as drug design and discovery. Additionally, graph-based[31-35] molecular representation learning methods show superiority compared to other molecular representations (i.e., molecular descriptors and fingerprints) due to the superior capabilities of graph neural networks (GNN) to model graph-structured data.

Molecules can be naturally depicted as graphs, with nodes and edges corresponding to atoms and chemical bonds. In 2015, Duvenaud et al.[36] first described a graph convolutional network (GCN) operator to create neural fingerprints, which achieved better performance in molecular property prediction tasks compared to the commonly used circular fingerprints. In 2017, Gilmer et al.[37] analyzed the similarities among several GNN models and proposed a universal framework termed Message Passing Neural Networks (MPNN) for the prediction of molecular properties, which consists of two phases, namely the message passing phase and readout phase. In 2019, Yang et al.[31] reported a directed MPNN (D-MPNN, also known as Chemprop), which updates information of directed bonds rather than atoms. In 2020, Xiong et al.[32] introduced an outstanding attention-based interpretable model (called Attentive FP) that attained well-performing results on various benchmark datasets for molecular property tasks. In 2021, Wu and coworkers proposed the HRGCN+ model that used molecular graphs and molecular descriptors to jointly predict molecular properties, achieving comparable or superior predictive performance.[34]

Despite these fruitful successes of GNN-based molecular property prediction, there are still several issues worth investigating. Firstly, the existing GNN architectures only take the spatial attention of atoms in the readout phase into account,[32–34] while the importance of atomic feature channels is usually ignored or rarely explored. Applying the attention mechanism to the feature channels is important because different atomic features may have distinct contributions to property prediction. In addition, compared with GNN models in CV and NLP,[38–40] mainstream GNN models neglect the hierarchical information of molecules[41] and lack prior knowledge of the domain. For instance, the role of molecular fragments in molecules can be analogous to patches in images or sentences in documents. In this sense, it is crucial to integrate knowledge from the chemistry domain into the GNN model for the prediction of molecular property.

To address the above-unresolved issues, in the present study, we proposed a well-designed hierarchical informative graph neural network (called HiGNN, Figure 1) framework for predicting the properties of molecules by exploiting both molecular graphs and fragments information simultaneously. Specifically, we first designed an interaction block between atoms in HiGNN framework, borrowing the principle of neural tensor networks for knowledge graph reasoning.[42] Meanwhile, a feature-wise attention (FA) block was proposed to recalibrate the atom representations after the message passing phase, which is capable of selectively focusing on salient features. Most importantly, we empirically used the BRICS (Breaking of Retrosynthetically Interesting Chemical Substructures),[43] a molecule decomposition algorithm based on chemistry domain knowledge, to cut molecules into fragments, and the attributed graph and fragments of the molecule were then input into the GNN encoder to obtain global and hierarchical representations in parallel. The performances of the HiGNN model were systematically evaluated on 11 real-world drug discovery-related benchmark datasets spanning physicochemical, biophysics, physiology, and toxicity. Comprehensive experiments illustrated that the HiGNN model achieved state-of-the-art (SOTA) performance on 10 of 14 learning tasks involving the 11 benchmark datasets compared to all baseline CML and DL models. Furthermore, extensive ablation studies demonstrated that both the BRICS and FA blocks can improve the prediction performance of the HiGNN model. Given that providing informative explanations remains a top priority for drug discovery tasks,[44] we therefore explored the interpretability of the HiGNN model through a molecule-fragment similarity mechanism, which may help chemists and pharmacists identify the key components of molecules for designing better molecules with desired properties or functions.

## METHODS

**Generic GNN.** Typically, a molecule can be depicted as an undirected graph $G = (V, E)$, where $V$ denotes the node (atom) set with $|V| = N$ nodes and $E \subseteq V \times V$ denotes the edge (bond) set with $M$ edges. Let every node $v_i \in V$ and edge $e_{ij} = (v_i, v_j) \in E$ have initial attributes $x_i \in R^{d_n}$ and $e_{ij} \in R^{d_e}$, where $d_n$ and $d_e$ refer to the feature dimensions of nodes and edges, respectively.

For molecular property prediction tasks, most GNN-based models follow a message-passing paradigm[37] based on three functions (i.e., message-passing function, aggregation function, and update function) to iteratively extract atomic features. The layer $k$ of the message passing phase can be formulated as follows:

$$m_i^{(k)} = Aggregate^{(k)}\big(\{Message^{(k)}\big(h_i^{(k-1)}, h_j^{(k-1)}, e_{ji}\big): j \in \mathcal{N}(i)\}\big), \tag{1}$$

$$h_i^{(k)} = Update^{(k)}\big(h_i^{(k-1)}, m_i^{(k)}\big), \tag{2}$$

where *Message*, *Aggregate*, and *Update* denote the message passing, aggregation and update functions, respectively. $h_i^{(k-1)}$ is the hidden state of node $i$ in the layer $(k-1)$ and $e_{ji}$ is the feature vector of edge from node $j$ to node $i$, and $\mathcal{N}(i)$ defines the neighborhood set of nodes $v$. Furthermore, a readout function is applied to form the whole graph-level representation $h_G$ according to

$$h_G = Readout(\{h_i^{(K)} | v_i \in G\}). \tag{3}$$

where $K$ refers to the final iteration, and *Readout* is a permutation invariant function that operates on the node set. Finally, we obtain the prediction $\hat{y} = f(h_G)$, where $f(\cdot)$ is a feed-forward neural network predictor.

**BRICS Fragment Representation.** Although graph-based DL approaches for molecular property prediction have been highly effective, little attention has been paid to the domain knowledge of chemistry, such as the hierarchical information of molecules. To this end, according to 16 cleavage rules (Figure 1d), the BRICS algorithm was used to obtain various fragments information through breaking the retrosynthetically relevant chemical bonds of the molecule.[43] It should be pointed out that not every molecule has a predefined chemical environment in BRICS algorithm, meaning that not all molecules can be cleaved into chemical fragments using the BRICS algorithm (Supporting Information Figure S1). The distribution of BRICS fragments for each dataset was counted and provided in Supporting Information Figures S2-S12.

Given a molecular graph $G = (V, E)$, the BRICS algorithm maps $G$ into a set of fragments with dummy atoms $S = \{S_1, S_2, \cdots, S_T\} = Brics(G)$, where *Brics* is implemented by the RDKit python package (https://www.rdkit.org/), and $T$ is the number of fragments. Then, we drop the dummy atom tagged on each fragment, thereby obtaining a new set of fragments $\tilde{S} = \{\tilde{S}_1, \tilde{S}_2, \cdots, \tilde{S}_T\}$, where $\tilde{S}_T$ is a subgraph of $G$ (if a molecule cannot be cleaved, we consider the molecule itself to be a fragment, in this case $T = 1$). Intuitively, the set $\tilde{S}$ can be described as the chemical fragment space of the molecules, which might encompass meaningful hierarchical information.

**The Architecture of HiGNN.** As shown in Figure 1c, the HiGNN architecture mainly consists of three modules: a GNN encoder for the molecular graph and its fragments, a dual-path combiner, and a predictor. As shown in Figure 1b, the initial atomic and bond features (Supporting Information Tables S1 and S2) assigned to the molecular graph and fragments are similar to the Attentive FP,[32] except that we additionally append pharmacophore and scaffold information to the atomic features. As shown in Figure 1e, we propose a novel GNN encoder for learning molecular graphs and fragments representations equipped with feature-wise attention (FA) block (Figure 1g). A dual-path combiner is then used to further extract the hierarchical information of molecules, which contains an attention-and-fusion block. Finally, the predictor module is used to produce the ultimate prediction.

**Interactive GNN Encoder.** In this article, we simply treat each atom as an entity and chemical bonds as the relationships between different entities. Accordingly, an intramolecular interaction layer is designed to capture the localized information. Figure 1a shows the differences in graph-based molecular representation learning among GCN,[45] graph attention networks (GAT),[46] and our interactive GNN encoder. We suppose that $h_i \in R^d$ is the hidden state of the target atom $i$ in the middle layer, as is $h_j \in R^d$ to the neighbor atom $j$. We first compute $h_i \leftarrow W h_i$, $h_j \leftarrow W h_j$ by a linear transformation, where $W \in R^{d \times d}$. Thus, as shown in Figure 1f, the multiple interaction scores $\alpha_{i,j} \in R^a$ between atom $i$ and $j$ can be defined as follows:

$$\alpha_{i,j} = tanh(h_i^T W_1^{[1:a]} h_j + W_2[h_i, h_j] + b), \tag{4}$$

where $a$ is a hyperparameter, $W_1^{[1:a]} \in R^{d \times d \times a}$, $W_2 \in R^{a \times 2d}$, and $b \in R^a$ are trainable parameters of the scoring function, and $h_i^T W_1^{[1:a]} h_j$ represents multiple bilinear tensor product that computes each entry by $h_i^T W_1^i h_j$ with slice $i = 1, \cdots, a$. We then reshape the neighbor information $h_j \in R^{a \times \bar{d}} \leftarrow Reshape(h_j, a, \bar{d})$, where $\bar{d} = \frac{d}{a}$ denotes subspace dimensions at different positions.[24] Once $\alpha_{i,j}$ and $h_j$ are obtained, the message passing phase operates according to the following *Message*, *Aggregate* (i.e. summation) functions

$$m_i = ReLU\left(\sum_{j \in \mathcal{N}(i)} Reshape(\alpha_{i,j} h_j, d)\right), \tag{5}$$

where $m_i \in R^d$ is the information aggregated from the neighboring atom(s), *ReLU* is a nonlinear activation function, and *Reshape* concisely refers to the concatenation of different subspaces.

In particular, we add bond features $e_{ji} \in R^d$ in case they are provided. Similar to the atomic features, we compute $e_{ji} \leftarrow W_e e_{ji}$ as well, where $W_e \in R^{d \times d}$. Hence, these functions are reformulated as follows:

$$\alpha_{i,j} = tanh(h_i^T W_1^{[1:a]} h_j + W_2[h_i, e_{ji}, h_j] + b), \tag{6}$$

where $W_2 \in R^{a \times 3d}$ is a learned matrix. Therefore, the model can extract edges information in an appropriate manner.

Afterwards, a gating mechanism is adopted to combine the hidden state $h_i$ and the message $m_i$. We concatenate $h_i$ and $m_i$ with the element-wise difference value $(h_i - m_i)$ and multiply it with a trainable weight $W_3 \in R^{3d \times d}$, i.e., $\beta = W_3^T[h_i, m_i, h_i - m_i]$. Thereby, the *Update* function is described as follows:

$$h_i^{'} = \beta \odot h_i + (1 - \beta) \odot m_i, \tag{7}$$

where $\odot$ denotes the Hadamard product and $h_i^{'}$ is the updated hidden state. We implement the gating block to share parameters (i.e., $W_3$) of each layer.

HiGNN simply implements the summation as the *Readout* function in the readout phase. Given the final updated atomic features set $\{h_1^{(K)}, h_2^{(k)}, \cdots, h_N^{(k)}\}$ after $k$ iterations of the message passing phase mentioned above, the representation of the molecule is calculated according to

$$h_G = \sum_{i=1}^{N} h_i^{(k)}, \tag{8}$$

Meanwhile, we also fed the fragment set $\tilde{S}$ into the GNN encoder. Notably, the encoders of the molecular graph and fragments share the same network structure and weights. A set of fragment representations is eventually obtained according to

$$S_G = \{s_1, s_2, \cdots, s_T\}. \tag{9}$$

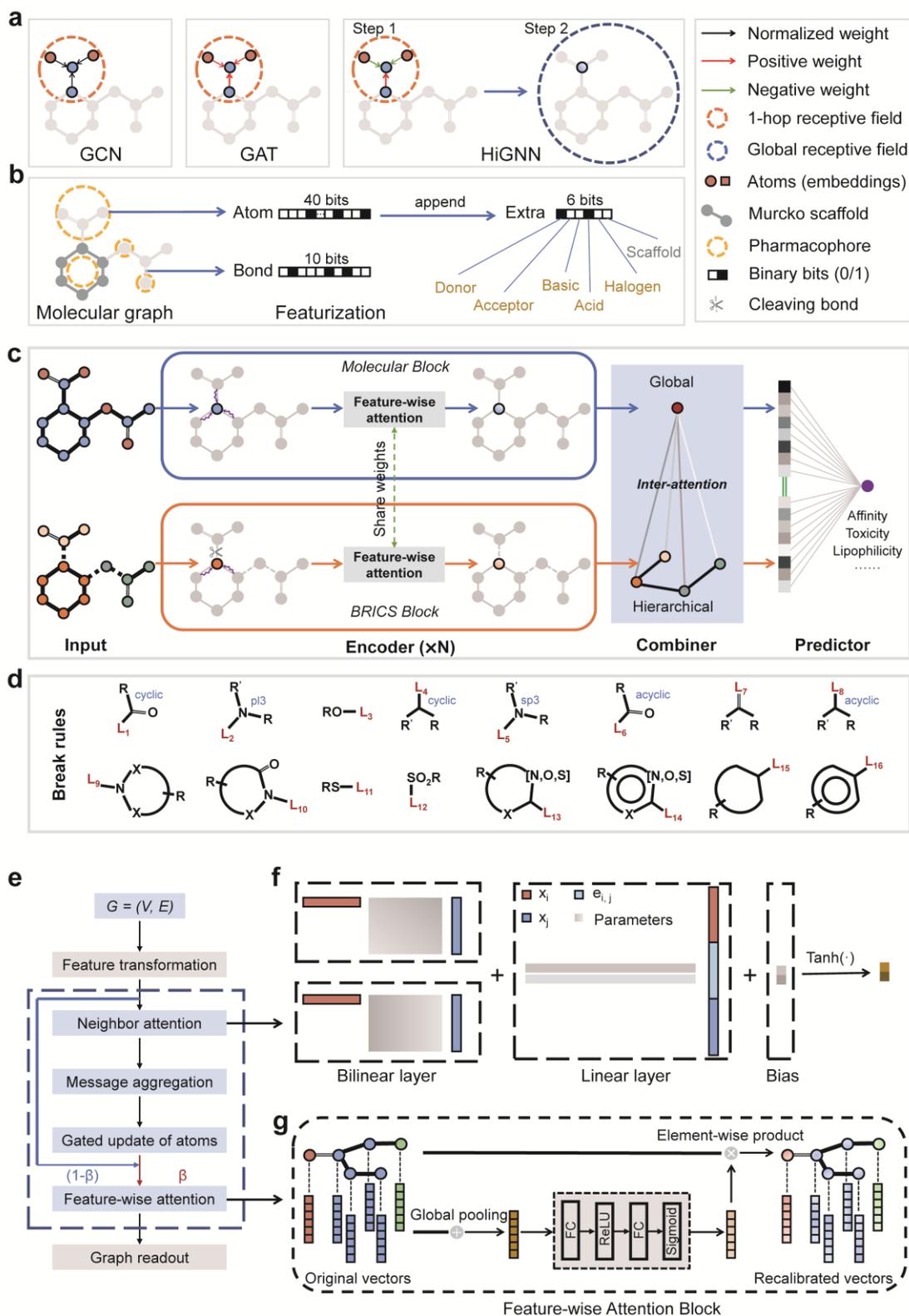

**Figure 1.** The overview of the HiGNN network architecture and its detailed elements. (a) The differences in graph-based molecular representations among GCN, GAT, and HiGNN. (b) The sketch of additional

pharmacophore and molecular scaffold features proposed in this study. (c) Overview of the HiGNN network architecture. (d) The cleavage rules according to different chemical environments defined in BRICS. (e) The workflow of the GNN encoder designed in the HiGNN framework. (f) Illustration of the intramolecular interaction layer. (g) The details of the feature-wise attention block.

**Feature-wise Attention Mechanism.** Herein, a feature-wise attention (FA) block, inspired by Hu et al. and Woo et al.,[47,48] is proposed to adaptively recalibrate the atom representations by modeling the relationships between feature dimensions. The detailed structure of the FA block is depicted in Figure 1g. Once the updated hidden state $h_i'$ in Equation (7) is obtained, we first squeeze the temporary global information into two different feature descriptors $f_{G-sum} = \sum_{i=1}^{N} h_i'$, $f_{G-max} = Max\{h_i' | v_i \in G\}$, which denote sum-pooling and max-pooling in feature channels, respectively. Then, we feed $f_{G-sum}$ and $f_{G-max}$ into a shared excitation operator to obtain the feature-wise weight $c \in R^d$ as follows:

$$c = \sigma(W_5 \delta(W_4 f_{G-sum}) + W_5 \delta(W_4 f_{G-max})), \quad (10)$$

where $W_4 \in R^{\frac{d}{r} \times d}$ and $W_5 \in R^{d \times \frac{d}{r}}$ are trainable parameters, and $\sigma$, $\delta$ refer to the Sigmoid and ReLU function, respectively. The reduction ratio r is utilized to limit model complexity and aid generalization.[47] Therefore, $h_i'$ is recalibrated according to

$$\widetilde{h_i'} = c \odot h_i'. \quad (11)$$

where $\odot$ is the Hadamard product, and $\widetilde{h_i'}$ is the recalibrated hidden state. Consequently, a set of recalibrated hidden states $\widetilde{H_G'} = \{\widetilde{h_1'}, \widetilde{h_2'}, \cdots, \widetilde{h_N'}\}$ is sent to the next layer. We also deploy the FA block that shares the parameters of each layer.

**Dual-path Combiner.** To integrate the co-representation of the molecule graph and its fragments, a dual-path combiner is introduced based on the following two main steps. First, an inter-attention strategy is adopted to grasp the intrinsic connections between the molecule and its fragments. Given the $h_G$ and $S_G = \{s_1, s_2, \cdots, s_T\}$ derived from the GNN encoder mentioned above, the attention scores on the fragments set are computed using the additive attention mechanism:

$$\alpha_t = softmax(LeakyReLU(a^\top[W_6 h_G || W_6 s_t])), \quad (12)$$

where $W_6 \in R^{d \times d}$ and $a \in R^{2d \times 1}$. The hierarchical information of fragments is then compacted into

$$s_G = \sum_{t=1}^{T} \alpha_t s_t. \quad (13)$$

Particularly, we apply multi-head attention with four heads to fully capture the relationships between the molecule and its fragments. Second, the $h_G$ and $s_G$ are concatenated and then passed through a predictor to make the final prediction $\hat{y} = f(h_G || s_G)$.

**Benchmark Datasets.** To benchmark the performance of the HiGNN model, 11 widely used drug discovery-related datasets were picked from Wu et al.,[49] covering physicochemical (ESOL, FreeSolv, and Lipo), biophysics (MUV, HIV, and BACE), physiology and toxicity (BBBP, Tox21, ToxCast, SIDER, and ClinTox). The number of tasks for these datasets ranges from 1 to 182, and the size of the datasets varies from 642 to 93,127. Details of the datasets were summarized in Supporting Information Table S3.

**Training Protocol and Hyperparameter Optimization.** HiGNN was implemented with the PyTorch[50] and PyTorch geometric library[51] and all training experiments were conducted on Tesla V100 GPUs. Parameters were updated using the Adam optimizer[52] and the entire network was trained in an end-to-end fashion. Following the protocol of Yang et al.,[31] 10 runs with different random seeds were utilized for randomly splitting all datasets into training, validation, and test sets, with a ratio of 8:1:1. Additionally, scaffold-based splitting was performed on BACE, BBBP, and HIV datasets. In accordance with the evaluation metrics of the reported CML and DL models, root-mean-square error (RMSE) and area under the receiver operating characteristic curve (ROC-AUC) were chosen as the evaluation metric for regression and classification tasks. Exceptionally, the MUV task was evaluated by the area under the precision-recall curve (PRC-AUC) recommended by Wu et al.,[49] since the positive and negative data points of this dataset are extremely imbalanced.

In this work, hyperparameters of each HiGNN model were optimized using the Tree Parzen Estimator algorithm implemented in the Hyperopt python package.[53] The following hyperparameters were selected during the training of the HiGNN model: the dimension of features, the depth of the GNN encoder, dropout rate, the number of slices in Equation (6), the reduction ratio in Equation (10), learning rate, and weight decay. Details of hyperparameters for each task were listed in Supporting Information Table S4.

## RESULTS AND DISCUSSION

**Performance Results of HiGNN on Benchmark Datasets.** In the present study, we evaluated the proposed HiGNN model on 11 commonly used and publicly available drug discovery-related datasets from Wu et al.,[49] including classification and regression tasks. According to previous studies, 14 learning tasks were designed based on 11 benchmark datasets, including 11 classification tasks based on random- and scaffold-splitting methods and three regression tasks based on the random-splitting method (Table 1). Commonly used DL and CML models, including SOTA graph-based DL methods (GCN, GAT, Attentive FP, Chemprop, and HRGCN+), as well as the advanced CML XGBoost model based on descriptors, were used as baseline models. The performance results of these baseline models were collected from the original papers.[31,34] The optimal hyperparameters of HiGNN models based on random-splitting and scaffold-splitting were summarized in Supporting Information Tables S5 and S6, respectively. In addition, for each dataset, the detailed results of the evaluation metric (i.e., ROC-AUC, PRC-AUC, or RMSE) of HiGNN under 10 random seeds were listed in Supporting Information Tables S7 and S8, and the average value was calculated to represent the final performance.

As shown in Table 1, HiGNN model achieved the best performance on 10 of 14 learning tasks, and performed second-ranked on the SIDER dataset when using the random-splitting. All these results demonstrate that HiGNN is capable of learning to accurately predict a broad range of molecular properties. Detailed analysis of the performance of HiGNN on three different types of datasets is described and discussed as follows:

Based on the long-term drug discovery practices, researchers have found that the success of candidate compounds in late-stage clinical trials is largely related to ADMET properties, thus it is of great importance to predict ADMET properties such as lipophilicity, water solubility, and hydration free energy. As shown in Table 1, HiGNN achieves the lowest RMSE value on each regression dataset, demonstrating the superiority of HiGNN on regression tasks. Compared with the second-ranked model, there are approximately 0.7%, 1.2%, and 5.5% improvements on the Lipo, FreeSolv, and ESOL datasets, respectively, indicating that HiGNN method can accurately predict ADMET-related properties of molecules in the early stages of drug discovery.

**Table 1. Predictive performance results of HiGNN on the drug discovery-related datasets.**

| Dataset | Split Type | Metric | Chemprop | GCN | GAT | Attentive FP | HRGCN+ | XGBoost | HiGNN |
|---|---|---|---|---|---|---|---|---|---|
| BACE | random | ROC-AUC | **0.898** | **0.898** | 0.886 | 0.876 | 0.891 | 0.889 | 0.890 |
| | scaffold | ROC-AUC | 0.857 | | | | | | **0.882** |
| HIV | random | ROC-AUC | 0.827 | **0.834** | 0.826 | 0.822 | 0.824 | 0.816 | 0.816 |
| | scaffold | ROC-AUC | 0.794 | | | | | | **0.802** |
| MUV | random | PRC-AUC | 0.053 | 0.061 | 0.057 | 0.038 | 0.082 | 0.068 | **0.186** |
| Tox21 | random | ROC-AUC | 0.854 | 0.836 | 0.835 | 0.852 | 0.848 | 0.836 | **0.856** |
| ToxCast | random | ROC-AUC | 0.764 | 0.770 | 0.768 | **0.794** | 0.793 | 0.774 | 0.781 |
| BBBP | random | ROC-AUC | 0.917 | 0.903 | 0.898 | 0.887 | 0.926 | 0.926 | **0.932** |
| | scaffold | ROC-AUC | 0.886 | | | | | | **0.927** |
| ClinTox | random | ROC-AUC | 0.897 | 0.895 | 0.888 | 0.904 | 0.899 | 0.911 | **0.930** |
| SIDER | random | ROC-AUC | **0.658** | 0.634 | 0.627 | 0.623 | 0.641 | 0.642 | 0.651 |
| FreeSolv | random | RMSE | 1.009 | 1.149 | 1.304 | 1.091 | 0.926 | 1.025 | **0.915** |
| ESOL | random | RMSE | 0.587 | 0.708 | 0.658 | 0.587 | 0.563 | 0.582 | **0.532** |
| Lipo | random | RMSE | 0.563 | 0.664 | 0.683 | 0.553 | 0.603 | 0.574 | **0.549** |

Each dataset was split using the corresponding data-splitting codes from published studies. The HiGNN models used the same dataset and data splitting method to fairly compare all baseline models including GCN, GAT, Chemprop, Attentive FP, HRGCN+ and XGBoost models. Bold font represents the model that outperforms all the other models. The performance results of GCN, GAT, Attentive FP, HRGCN+ and XGBoost models were collected from Wu et al., while the optimized results of the Chemprop models were from the original study.[31,34] D-MPNN (Chemprop): directed message passing neural networks; GCN: graph convolutional networks; GAT: graph attention networks; XGoost: extreme gradient boosting.

Bioactivity screening is a key part of modern drug development, where the primary goal is to identify, from a large pool of compounds, molecules that are active against targets relevant to diseases. Therefore, enhancing ligand-based virtual screening with DL can help screen compounds with higher success rates in the early stages of drug discovery. Table 1 indicates that HiGNN shows significant advantage on the MUV dataset with the highest PRC-AUC value of 0.186. GCN and Chemprop performed the best on the BACE dataset when using random-splitting method, followed by HRGCN+, HiGNN, XGBoost, GAT, and Attentive FP. Meanwhile, GCN and Chemprop also showed considerable predictive performance on the HIV dataset. Scaffold hopping is a fundamental task in drug discovery, which aims to find compounds with similar activities but containing different core structures. Compared to random-splitting, scaffold-splitting is more practical and challenging for DL models. Table 1 illustrates that HiGNN performs the best on both BACE and HIV datasets when using the scaffold-splitting method, suggesting that our model has excellent scaffold hopping ability. Overall, HiGNN achieved the best performance on three of five learning tasks for these three biophysics datasets (BACE, HIV, and MUV), implying that HiGNN exhibits comparable or superior competitiveness in the tasks of molecular pharmacological activity prediction.

The macroscopic effects of drugs on the human body, such as blood-brain barrier permeability (BBBP), side effects, and toxicities, will directly affect the success of new drug development. Accurate prediction of these physiological and toxicological properties can help reduce costs, shorten cycles, and improve success rates for drug development. As shown in Table 1, HiGNN outperformed all baseline models on the BBBP (random-splitting and scaffold-splitting methods), ClinTox, and Tox21 datasets. HiGNN performed the second-ranked on the SIDER dataset with ROC-AUC value of 0.651, which is only slightly weaker than Chemorop (ROC-AUC = 0.658), and far better than the other models (GCN, GAT, Attentive FP, HRGCN+, and XGBoost). In addition, it also performed the third-ranked on the ToxCast dataset. All these results demonstrate that HiGNN can precisely predict the physiological and toxicological properties of compounds, which can be used to rule out improper molecules in the early stages of drug discovery.

**Table 2. Ablation study results on the variants of HiGNN.**

| Method | BACE | BBBP | ClinTox | SIDER | Tox21 | FreeSolv | ESOL | Lipo |
|---|---|---|---|---|---|---|---|---|
| HiGNN | **0.890** | **0.932** | **0.930** | 0.651 | **0.856** | **0.915** | **0.532** | **0.549** |
| w/o HI | 0.887 | 0.930 | 0.926 | **0.654** | 0.852 | 0.941 | 0.536 | 0.575 |
| w/o FA | 0.880 | 0.927 | 0.927 | 0.649 | 0.840 | 0.964 | 0.616 | 0.591 |
| w/o All | 0.877 | 0.918 | 0.924 | 0.639 | 0.830 | 0.946 | 0.670 | 0.623 |

Bold font represents the best-performing model. w/o HI represents HiGNN without hierarchical information. w/o FA represents that HiGNN without feature-wise attention. w/o All represents that HiGNN without hierarchical information and feature-wise attention. HI: Hierarchical information; FA: Feature-wise attention.

**Effectiveness of Hierarchical Information and Feature-wise Attention.** As shown in Figure 1c, HiGNN is a dual-path framework that combines global and hierarchical information of molecules with the FA block, which is completely different from the existing GNN models. To further investigate the effectiveness of two well-designed components in the HiGNN architecture, namely BRICS and FA blocks, we conducted ablation studies to decouple our networks. Concretely, we employed three variants of HiGNN as follows:

1. HiGNN *without hierarchical information* (w/o HI): removing the BRICS block in Figure 1c.
2. HiGNN *without feature-wise attention* (w/o FA): removing the FA block in Figure 1c.
3. HiGNN *without hierarchical information and feature-wise attention* (w/o All): removing both the BRICS and FA blocks in Figure 1c.

The performances of all variants of HiGNN were conducted on eight datasets (BACE, BBBP, ClinTox, SIDER, Tox21, FreeSolv, ESOL, and Lipo), and the detailed results were presented in Table 2. For a fair comparison, the experimental setup of these variants is consistent with the HiGNN.

Firstly, we decoupled our HiGNN into two single-path networks, the upper path (i.e., w/o HI) and the lower path (Figure 1c). As shown in Table 2, we found that (1) utilizing only the lower path yielded unsatisfactory results, indicating that BRICS-based fragment information alone is insufficient as a molecular representation; and (2) HiGNN achieves better results in seven out of eight datasets compared to its variant (w/o HI), demonstrating that BRICS-based fragment information encoded in the co-representation could help to accurately predict molecular properties. To visualize the learned representation of molecules, we used the latent vectors, $h_G$ and $s_G$, obtained from the readout phase to characterize the global and hierarchical information of the molecules, respectively. Taking the BBBP dataset as an example, Figure 2a shows the permeable (red) and impermeable molecules (blue) in the test set by projecting the learned representation using the t-distributed stochastic neighbor embedding (t-SNE). In both the molecular-level (Figure 2a, left) and fragment-level (Figure 2a, right) spaces, molecules are roughly divided into two clusters with different permeability, confirming that our HiGNN model could extract the meaningful latent representations concerning the property. For one thing, HiGNN can separate the clusters better in the molecular-level space than in the fragment-level space. For another, two molecules with different permeability (Figure 2a, middle) overlap in the molecular-level space but are separated in the fragment-level space because of discrepancies in their molecular fragment information. Based on these observations, we infer that HiGNN as a fragment-assisted network can boost the predictive performance via absorbing the hierarchical information.

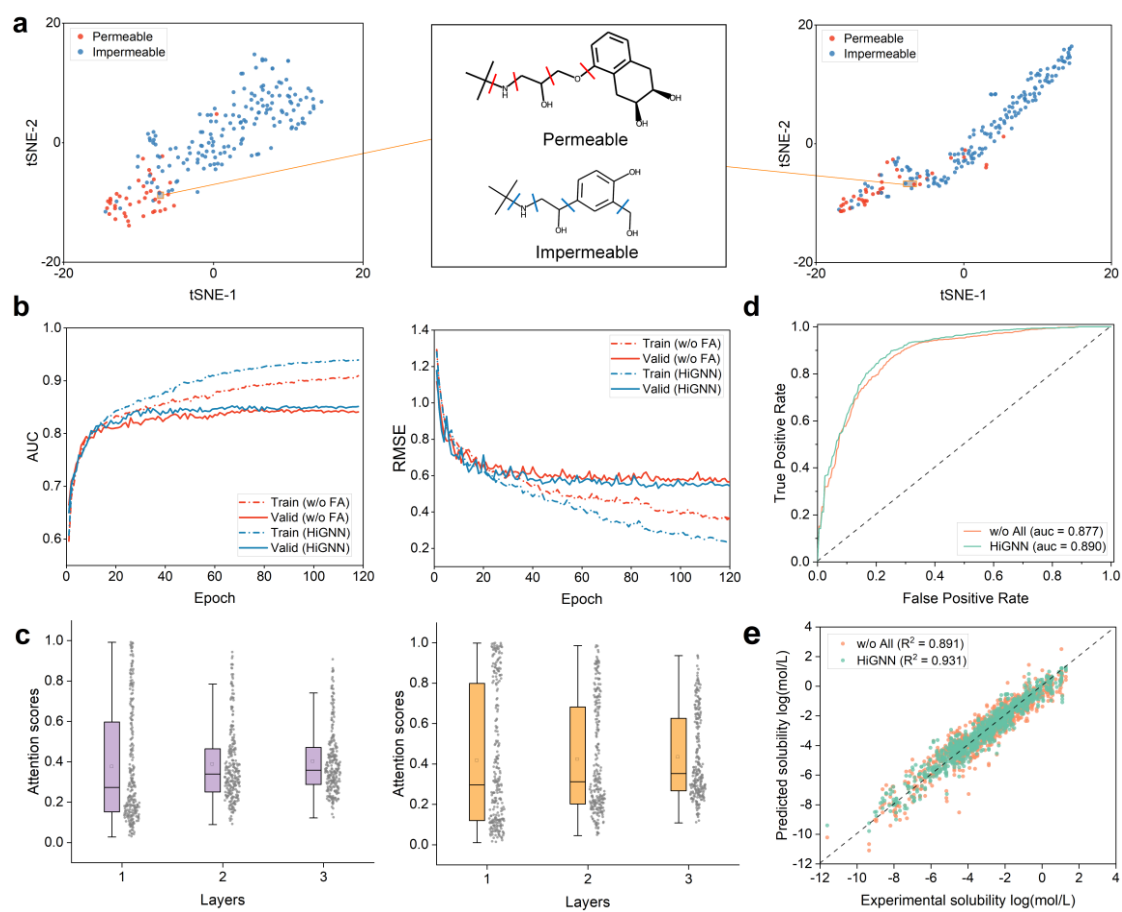

**Figure 2.** Investigation of the impact on the BRICS and FA blocks. (a) Visualization of the test set on the BBBP dataset using t-SNE. The axes are the top two important components of the learned representation of molecules by HiGNN's two paths, i.e., molecular-level (left) and fragment-level space (right). Meanwhile, two molecules with different permeability are zoomed in to display the molecular structure (middle). (b) The learning curves of HiGNN and its variant (w/o FA) on the Tox21 (left) and Lipo (right) datasets. (c) The distributions of 256-dimensions attention scores generated by FA blocks at different layers on the Tox21 (left) and Lipo (right) datasets. (d) The average receiver operating characteristic (ROC) responses on the BACE dataset with 10 random runs by HiGNN and its variant (w/o All). (e) The average of predicted results on the BACE dataset of 10 random runs by HiGNN and its variant (w/o All).

Secondly, we probed the impact of the FA block in HiGNN on the molecular property prediction tasks. Table 2 illustrates that the variant (w/o FA) drops significantly in predictive performance on all datasets, validating the superiority of our well-designed FA block. There are two benefits of the FA block. On the one hand, the module can adaptively recalibrate atomic representations after the message passing phase by modeling the relationships between feature dimensions. On the other hand, the module can implicitly enlarge the receptive field of GNN layers, since this process squeezes all atomic information. To gain insight into the training process of our HiGNN and its variant (w/o FA), we plotted the training curves of two models running on the Tox21 and Lipo datasets. As depicted in Figure 2b, in both classification and regression tasks, HiGNN offers sustained improvements on the training and validation sets throughout the training process, indicating that the FA block leads to a finer optimization route. To elucidate the function of the FA block, we further analyzed the attention scores after the Sigmoid activation. Figure 2c illustrates the distribution of attention scores in individual feature dimensions generated by the FA block at different layers. It is worth noting that the distribution in each layer shows significantly distinct characteristics, suggesting that the FA block can be useful to adaptively capture layer-wise information. Furthermore, the attention scores at the first layer are mostly close to 0, indicating that some insignificant information would be filtered out at the early stages due to the inclusion of FA block, that is, the HiGNN model can selectively focus on the salient features. Additionally, as the number of layers deepens, the average value of the attention score increases and the overall distribution becomes more concentrated, proving that the atomic features are well-calibrated at the later stages. The role of the reduction ratio $r$ was also assessed in 1, 2, 4, 8, and 16 (Supporting Information Table S9). Experimental results show that the optimal reduction ratio varies for distinct datasets. Collectively, the FA block enables the HiGNN model with more expressive power compared to its variant (w/o FA) architecture.

Finally, to compare with HiGNN and its variant (w/o All), we visualized metrics for both classification and regression tasks to gain further insight into the advantage of HiGNN. Taking the HiGNN and its variant (w/o All) on the BACE dataset as an example, Figure 2d illustrates that the HiGNN (green) curve is mostly above its variant (red) curve, suggesting that HiGNN provides better classification performance. Meanwhile, Figure 2e displays the relevance of predicted results and experimental results on the ESOL dataset based on HiGNN and its variant (w/o All). Apparently, HiGNN presents better predictive regression performance with a higher $R^2$ value of 0.931. These outcomes verify that the entire HiGNN framework strongly dominates the simpler variant (w/o All).

**Analysis of featurization and convolution on GNN.** In this section, we presented a comprehensive analysis of HiGNN in terms of featurization and convolutional operation. Specifically, we aimed at evaluating the following three questions:

Do extra atomic features benefit HiGNN? In this work, extra pharmacophore (donor, acceptor, basic, acid, and halogen) and scaffold information (Figure 1b) were added to enrich the atomic features. To assess the effectiveness of the additional features and provide a fair comparison with Attentive FP, we conducted experiments on the eight datasets. As shown in Figure 3, HiGNN_46 (with additional features) produces steady gains on seven out of the eight datasets compared to HiGNN_40 (without additional features). Meanwhile, we could observe that HiGNN_40 is slightly inferior to the performance of Attentive FP on the Lipo dataset, while exceeding Attentive FP on the other seven datasets, indicating that HiGNN can achieve impressive performance using the same featurization scheme. Surprisingly, the addition of the extra features resulted in a large drop in ROC-AUC value on the BACE dataset, indicating that the selection of initialization features for molecular graphs is task-specific, and we leave this exploration to future work.

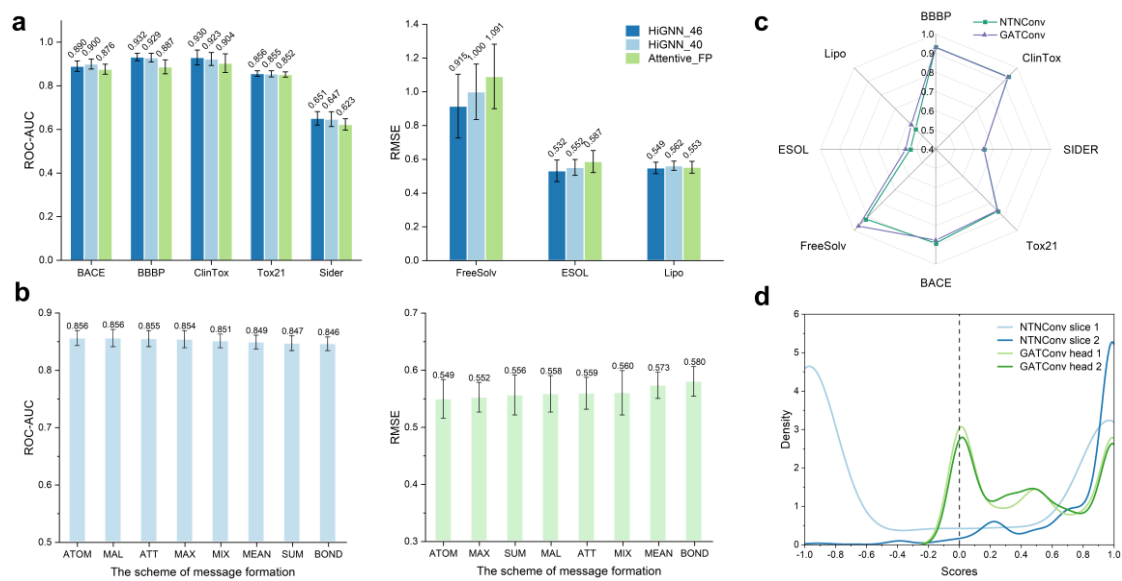

**Figure. 3** The analysis of featurization and convolution in HiGNN network architecture. (a) Comparison of performance results of HiGNN_40, HiGNN_46, and Attentive FP on eight datasets. HiGNN_46 represents that the initialization atomic features are 46-bits, including additional pharmacophore (donor, acceptor, basic, acid, and halogen) and molecular scaffold information. HiGNN_40 represents that the initialization atomic features are 40-bits, and the featurization method is the same as attentive FP. (b) The performance results of HiGNN using different types of message formation functions on the Tox21 (left) and Lipo (right) datasets. "ATOM" and "BOND" denote that the message is constructed by the neighboring atomic features and the connected bond features, respectively. "ATT" means removing bond information when calculating the interaction scores, that is, using Equation (4) instead of Equation (6). "MAX", "MEAN", and "SUM" represent using max, average, and sum pooling of atomic and bond features to create the message. "MUL" and "MIX" indicate the use of $h_j \cdot e_{ji}$ and $h_j + e_{ji} + h_j \cdot e_{ji}$ in the message-passing function, respectively. (c) Comparison performance results of HiGNN with NTNConv and GATConv on eight datasets. ROC-AUC and RMSE are used for evaluating the classification and regression tasks, respectively. (d) Distribution of edge coefficients for 50 random samples from the Tox21 test set. Note that the curve with coefficients below 0 is automatically fitted for GATConv, and in fact, the coefficients of this convolution method should be between 0 and 1, which is caused by kernel smooth using OriginPro Learning Edition.

Do chemical bond features matter? In the present study, we also take the properties of edges into account in the HiGNN model. Different types of message formation schemes were adopted to comprehensively study the effect of bond features, i.e., various communications of atomic features and bond features are incorporated into Equation (5). We performed experiments on the Tox21 and Lipo datasets to shed light on the role of bond features, and the results are illustrated in Figure 3b. The first point we want to mention is the comparison between "ATOM" and "BOND" schemes, which utilize only atomic and bond information, respectively. As shown in Figure 3b, the latter ("BOND" schemes) shows a large performance degradation on both tasks, reflecting that atoms are more informative than bonds. Furthermore, the scheme of "BOND" also achieves satisfactory results, implying that the incremental and layer-wise aggregation of bond features can extract some effective patterns as well. Secondly, we focus on the calculation of interaction scores in Equation (4), and the "ATT" scheme eliminates the participation of bond information. Compared to the "ATOM" scheme, this elimination led to a slight drop in ROC-AUC on the Tox21 dataset and a moderate rise in RMSE on the Lipo dataset, indicating the importance of bond features participating in the calculation. Driven by these insights, we additionally implemented several schemes, marked as "MAX", "MEAN", "SUM", "MAL", and "MIX". As shown in Figure 3b, the "MUL" scheme performed on par with the "ATOM" scheme on the Tox21 dataset, and the "MAX" scheme achieved comparable predictive power to the "ATOM" scheme on the Lipo dataset. Although these schemes presented different rankings in the two datasets, the "ATOM" scheme generally prevailed over other schemes in both tasks.

How does HiGNN's convolution differ from GAT? As mentioned above, we proposed an interactive convolution layer (termed NTNConv). In comparison with the original GAT convolution layer (termed GATConv),[46] we conducted experiments on eight datasets, and the results are shown in Figure 3c. Clearly, the proposed NTNConv in our study not only demonstrates a rival or even superior prediction accuracy to the GATConv on the classification tasks, but also achieves an average improvement of 5.0% on the regression tasks. It should be pointed out that the GATConv utilizes the Sigmoid activation and SoftMax normalization to estimate the attention coefficients, whereas the NTNConv utilizes additional bilinear layer and the Tanh activation (Figure 1f) to capture the interactions between atoms. Based on this difference, we randomly sampled 50 molecules from the Tox21 dataset and then visualize the edge coefficients $\alpha_{i,j}$, extracted from the last layer of HiGNN with NTNConv and GATConv, respectively. Figure 3d illustrates the distribution of the internal $\alpha_{i,j}$ of these molecules, and it can be found that the two convolutional layers present completely different distributions, revealing that the learning paradigms of these two convolutional layers are different. Furthermore, we conducted ablation study in terms of the depth in the GNN encoder, Supporting Information Figure S13 shows that NTNConv achieves better performance than GATConv at most layers, indicating that our HiGNN framework is capable of alleviating over-smoothing problem.

**The interpretation of HiGNN.** In this work, we analyzed the interpretation of HiGNN at the subgraph level through a molecular-fragment similarity mechanism. As delineated above, our inter-attention block utilizes four attention heads to capture the complex relationships between the molecule and its fragments, which motivates the representations of the molecule and fragments to be correlated with each other. As depicted in Figure 4a, we provide a simple method to determine the correlations by calculating the cosine similarity between the molecule and its fragments, which may assist in identifying the key fragments that are synthetically accessible. Taking the BACE and BBBP datasets as examples, we investigated whether HiGNN can capture critical patterns associated with bioactivity and permeability properties, respectively.

β-Secretase 1 (BACE-1), a crucial enzyme involved in the formation of amyloid β-peptides found in extracellular amyloid plaques of Alzheimer's disease (AD), is considered as an attractive target for the treatment of AD.[54] Herein, the optimal model architecture was picked to train on the BACE dataset using the scaffold-splitting method. After training, two molecules (identified as BACE_709 and BACE_1263,

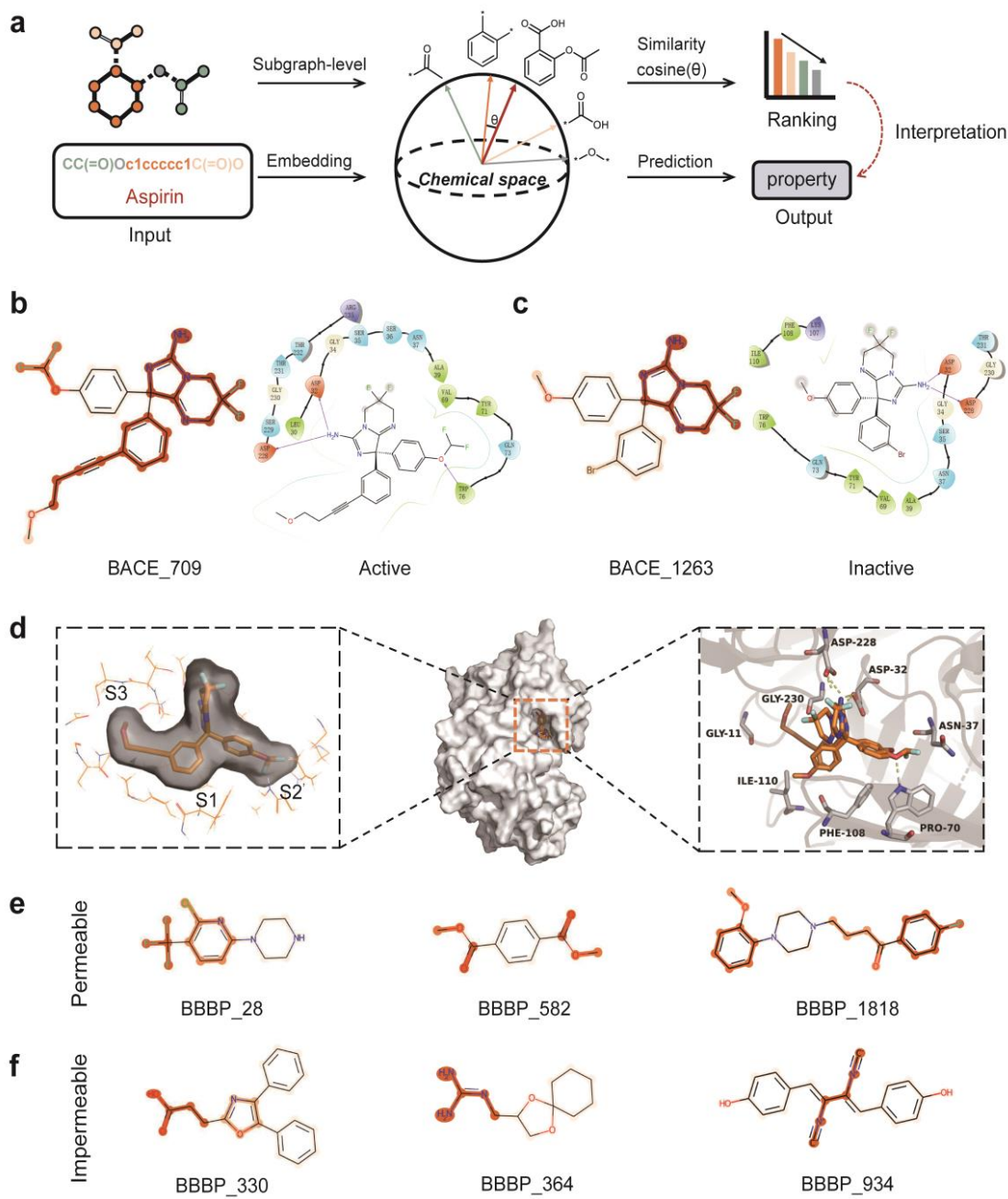

**Figure. 4** Interpretation of HiGNN on BACE and BBBP datasets. (a) Illustration of the molecular-fragment similarity mechanism. (b), (c) Demonstration of the colored active and inactive molecules against BACE-1, and the 2D protein-ligand interaction binding modes generated by using Glide SP docking. (d) Representation of the binding pocket (left); the solid surface of protein and ligand binding sites (middle); the predicted 3D binding modes of BACE_709 and BACE_1263 to BACE-1 (right). All figures were generated using PyMOL software (https://pymol.org/2/). (e), (f) Molecules (three permeable and three impermeable) with respect to the permeability predicted by the HiGNN model. The fragment-molecule cosine similarities are used to color the molecules.

Figures 4b, 4c) were chosen from the test set for case studies. Interestingly, these two molecules share the same scaffold, but BACE_709 ($pIC_{50}$ = 7.25) is active while BACE_1263 ($pIC_{50}$ = 5.47) is inactive.[55]

Based on this observation, our HiGNN model can highlight the attractive fragments of the two molecules that are in line with the cosine similarity scores (Figures 4b, 4c). Meanwhile, the binding modes of BACE_709 and BACE_1263 to BACE-1 were explored using Glide SP docking, and the 2D protein-ligand interactions of the two molecules are presented in Figures 4b, 4c, respectively. Glide scores suggested that BACE_709 (docking score = -7.33 Kcal/mol) showed better inhibitory effect than BACE_1263 (docking score = -4.26 Kcal/mol) against BACE-1, which was consistent with the experimental results. We can observe that the highlighted part of the molecule is consistent with the formed hydrogen bond interactions of the docking results, indicating that our model can automatically learn the key information from the well-annotated datasets. Furthermore, Figure 4d displays the detailed 3D binding patterns of two molecules against BACE-1. It is worth noting that an alkyne moiety of BACE_709 extends towards the S3 pocket, resulting in strong hydrophobic interactions, which is also in accordance with the highlighted part of BACE_709. However, these hydrophobic interactions were not observed in BACE_1263 due to the lack of the alkyne moiety, which may account for its poor inhibitory activity against BACE-1. Therefore, our model is capable of detecting the key patterns related to bioactivity interactions.

A major factor in the success of drugs acting on the central nervous system is the blood-brain barrier (BBB), which blocks most drugs, hormones, and neurotransmitters.[56] The blood-brain barrier permeability (BBBP) is related to the molecular weight, lipophilicity, dissociation constant, and other physicochemical properties of the drug. Similarly, we selected the optimal model architecture to train on the BBBP dataset using the random-splitting method. To observe the patterns learned by the HiGNN model, we colored molecules according to the cosine similarity scores. Figure 4e and Figure 4f display permeable and impermeable molecules sampled from the test set, respectively. For permeable molecules, we found that the HiGNN model pays more attention to the lipophilic groups, such as trifluoromethyl (BBBP_28), ester group (BBBP_582), and fluorophenyl (BBBP_1818). Conversely, for permeable molecules, the HiGNN model focuses more on hydrophilic group (BBBP_934). Meanwhile, molecules containing carboxyl groups such as BBBP_30 easily exist in the form of ions under the pH conditions of the body, while molecules containing guanidine groups such as BBBP_364 are prone to hydrolysis, indicating that they are difficult to penetrate the BBB. These visualizations demonstrate that our model can identify key groups in an accurate and interpretable manner, which is consistent with the perception of chemists.

CONCLUSION

In this study, we proposed HiGNN, a fragment-assisted and feature-wise attentive framework for molecular property prediction, which couples the molecular graph and fragment-level information of molecules in parallel. Specifically, two main strategies are introduced into the well-designed model: (1) a molecule decomposition algorithm named BRICS is incorporated into GNN to absorb fragment information; and (2) a plug-and-play feature-wise attention block to adaptively recalibrate atomic features after the message passing phase. Extensive experiments illustrate the superiority of HiGNN on a broad variety of challenging molecular property benchmarks compared to multiple competing baseline models. Meanwhile, comprehensive experiments demonstrated the effectiveness of both strategies. In addition, we provided a thorough analysis of featurization and convolution on GNN, which proves the advantage of our network and may lay the groundwork for future work. More importantly, the interpretation of HiGNN at the subgraph level is easily parsed and understood via a molecular-fragment similarity mechanism, indicating that our HiGNN as a powerful tool can help chemists and pharmacists identify the key components of molecules for designing better molecules with desired properties or functions. Collectively, we anticipate that HiGNN as a new easy-to-use and efficient AIDD tool can be used for drug discovery-related tasks.

## DATA AVAILABILITY

The datasets used in this study and the source code for HiGNN are publicly available at https://github.com/idrugLab/hignn.

## ASSOCIATED CONTENT

### Supporting Information

The Supporting Information is available free of charge via the Internet at http://pubs.acs.org.

## AUTHOR INFORMATION

### Corresponding Author

*Corresponding author: Ling Wang, School of Biology and Biological Engineering, South China University of Technology, Guangzhou 510006, China. E-mail: lingwang@scut.edu.cn.

### Author Contributions

All authors have given approval to the final version of the manuscript.

### Funding Sources

This work was supported in part by the National Natural Science Foundation of China (81973241) and the Natural Science Foundation of Guangdong Province (2020A1515010548).

### Notes

The authors declare no competing financial interest.

## ACKNOWLEDGMENTS

The authors thank the members of the idrugLab@SCUT, and we acknowledge the allocation time from the SCUTGrid at South China University of Technology.

## ABBREVIATIONS

GNN, graph neural networks; HiGNN, hierarchical informative graph neural networks; ADMET, absorption, distribution, metabolism, excretion, and toxicity; HTS, high-throughput screening; CADD, computer-aided drug design; QSAR, quantitative structure-activity relationships; AI, artificial intelligence; AIDD, AI-aided drug design; SMILES, simplified molecular-input line-entry systems; CML, conventional machine learning; SVM, support vector machine; RF, random forest; XGBoost, extreme gradient boosting; DL, deep learning; CV, computer vision; NLP, natural language processing; DNN, deep neural networks; CNN, convolutional neural networks; RNN, recurrent neural networks; GCN, graph convolutional networks; GAT, graph attention networks; MPNN, message passing neural networks; D-MPNN, directed message passing neural networks; FA, feature-wise attention; BRICS, breaking of retrosynthetically interesting chemical substructures; SOTA, state-of-the-art; RMSE, root-mean-square error; ROC-AUC, area under the receiver operating characteristic curve; PRC-AUC, area under the precision-recall curve; BACE-1, β-Secretase 1; AD, Alzheimer's disease; BBB, blood-brain barrier; BBBP, blood-brain barrier permeability

Graphic abstract

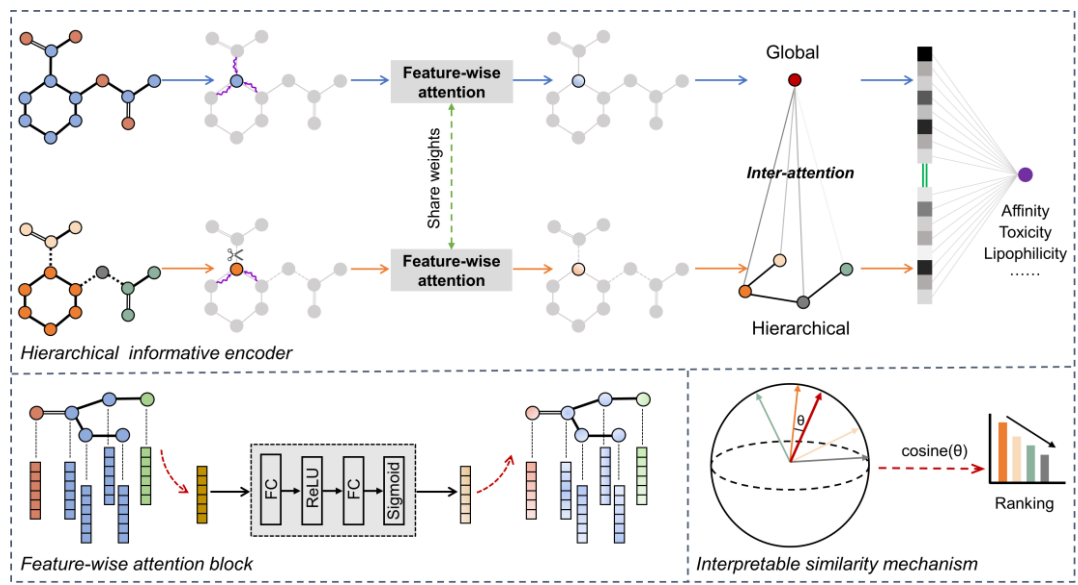